\documentclass{article}

\usepackage{arxiv}

\usepackage[utf8]{inputenc} 
\usepackage[T1]{fontenc}    
\usepackage{hyperref}       
\usepackage{url}            
\usepackage{booktabs}       
\usepackage{amsfonts}       
\usepackage{nicefrac}       
\usepackage{microtype}      
\usepackage{lipsum}		
\usepackage{graphicx}
\usepackage{natbib}
\usepackage{doi}
\usepackage{verbatim}
\usepackage{subcaption}
\usepackage[misc]{ifsym}
\usepackage{outlines}
\usepackage{algorithm}
\usepackage[noend]{algpseudocode}
\usepackage{amsmath}
\usepackage{array}
\usepackage{multirow}
\usepackage{subcaption}
\usepackage{listings}

\usepackage{pgf}
\usepackage{tikz}
\usetikzlibrary{arrows,automata}
\usetikzlibrary{positioning}

\tikzset{
    state/.style={
           rectangle,
           rounded corners,
           draw=black, very thick,
           minimum height=2em,
           minimum width=2em,
           inner sep=2pt,
           text centered,
           },
}
\usepackage{placeins}
\usepackage{amssymb}
\usepackage{enumitem}
\usepackage{pifont}

\title{Consumer Transactions Simulation through Generative Adversarial Networks under Stocks Constraints in Large-Scale Retail}

\author{ 
    \href{https://orcid.org/0000-0002-3434-6320}{\includegraphics[scale=0.06]{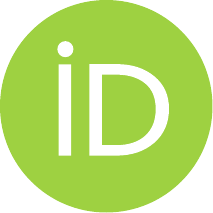}\hspace{1mm}Sergiy Tkachuk$^1$},
\href{https://orcid.org/0000-0001-6716-610X}{\includegraphics[scale=0.06]{orcid.pdf}\hspace{1mm}Szymon {\L}ukasik$^{1,3}$},
\href{https://orcid.org/0000-0002-3407-7570}{\includegraphics[scale=0.06]{orcid.pdf}\hspace{1mm}Anna Wr{\'o}blewska$^2$}\\
 	$^1$Systems Research Institute,
		Polish Academy of Sciences\\
		ul.\ Newelska 6, 01-447 Warsaw, Poland\\
		Email: \texttt{\{stkachuk,slukasik\}@ibspan.waw.pl} \\
 	$^2$Faculty of Mathematics and Information Science, Warsaw University of Technology, Warsaw, Poland \\ Email: \texttt{anna.wroblewska1@pw.edu.pl} \\
	$^3$Faculty of Physics and Applied Computer Science,
		AGH University of Science and Technology\\
		al.\ Mickiewicza 30, 30-059 Krak\'{o}w, Poland\\
		Email: \texttt{slukasik@agh.edu.pl}
 }

\date{}

\renewcommand{\shorttitle}{Consumer Transactions Simulation through Generative Adversarial Networks}

\hypersetup{
pdftitle={Consumer Transactions Simulation through Generative Adversarial Networks under Stocks Constraints in Large-Scale Retail},
pdfsubject={cs.LG},
pdfauthor={Sergiy Tkachuk, Szymon {\L}ukasik, Anna Wr{\'o}blewska},
pdfkeywords={Generative Adversarial Networks, Deep Learning, Transaction Embedding Representation, Consumer Behavior Modeling.},
}

\begin{document}

\maketitle

\begin{abstract}
In the rapidly evolving domain of large-scale retail data systems, envisioning and simulating future consumer transactions has become a crucial area of interest. It offers significant potential to fortify demand forecasting and fine-tune inventory management. This paper presents an innovative application of Generative Adversarial Networks (GANs) to generate synthetic retail transaction data, specifically focusing on a novel system architecture that combines consumer behavior modeling with stock-keeping unit (SKU) availability constraints to address real-world assortment optimization challenges. We diverge from conventional methodologies by integrating SKU data into our GAN architecture and using more sophisticated embedding methods (e.g., hyper-graphs). This design choice enables our system to generate not only simulated consumer purchase behaviors but also reflects the dynamic interplay between consumer behavior and SKU availability -- an aspect often overlooked, among others, because of data scarcity in legacy retail simulation models. Our GAN model generates transactions under stock constraints, pioneering a resourceful experimental system with practical implications for real-world retail operation and strategy. Preliminary results demonstrate enhanced realism in simulated transactions measured by comparing generated items with real ones using methods employed earlier in related studies. This underscores the potential for more accurate predictive modeling.
\end{abstract}


\section{Introduction} \label{intro}

Assortment planning, a cornerstone of retail operations~\cite{bahng2018assortment, kok2015assortment} for many large-scale chains, is fraught with complexities~\cite{rooderkerk2019omnichannel}. Determining the optimal mix of products a retailer should offer involves navigating a myriad of constraints, from limited display space to procurement budgets and strategic vendor decisions. The foundation for these decisions lies in merchandise categories, groups of stock-keeping units (SKUs)\footnote{A unique code or identifier assigned to a specific product to track inventory, sales, and manage stock levels} perceived as substitutes by customers~\cite{goyal2016near}. These decisions span from long-term strategic choices, such as defining merchandise variety, to tactical considerations like inventory levels and pricing~\cite{mendez2010branch}.

In the complex and dynamic field of retail, the ability to accurately model and understand customer behavior is crucial for optimizing operations and enhancing the customer experience, which further complicates the landscape~\cite{nazir2023exploring}. While some consumers enter a store with a specific product in mind, others are influenced by the retailer's displayed SKUs, ads, and discounts. Thus, a retailer's sales volume becomes a function of demand, customer preferences, and strategic and tactical decisions~\cite{gallego2019revenue}. Despite the wealth of consumer transaction data, many retailers still lean heavily on qualitative judgments. However, the tide is turning, with the industry gradually embracing a more data-driven approach~\cite{muller2020data}.

\subsection{Problem Statement}
Understanding and simulating customer behavior is paramount for optimizing operations and enhancing customer experience in the contemporary large-scale retail landscape. Companies gather business-critical data in consumer relationship management (CRM)~\cite{anshari2019customer} and enterprise resource planning (ERP)~\cite{adjie2020critical} systems to ensure smooth operations, better compliance, and effective data governance. Effectively modeling collected data for tangible commercial impact presents several challenges we aim to address with our research: data scarcity, versatility, and privacy.
    
    \textbf{Data Scarcity and Versatility:} While vast amounts of transactional data are available through loyalty programs, more than half of consumers can be inactive members calling out reward programs and irrelevant communication as primary reasons for inactivity~\cite{alshurideh2020loyalty}. This creates a fundamental difficulty in consumer behavior analysis as collected data cannot be generalized through repeated purchase patterns and limits to the most loyal customers, enabling their identification via point-of-sales transactional systems. One of the recent studies also finds that Gen Z~\cite{genz} are least loyal to retailers and tend to choose them more wisely than any other preceding generations~\cite{thangavel2022consumer}, which also sets an additional challenge for retails in the long-term and enforces to focus substantially on more advanced and out-of-the-box analytical methods.
    
    \textbf{Privacy Considerations:} As retailers increasingly rely on data-driven strategies to optimize operations and customer experience, data privacy's ethical and legal implications become more pronounced. Collecting and analyzing consumer data, especially through CRM and ERP systems~\cite{stankov2020vulnerability}, often involve sensitive information such as personal identifiers, purchase history, and payment details. While this data is invaluable for analytics and personalization, it poses significant risks if mishandled or subjected to unauthorized access. According to a recent study on data privacy in the era of big data~\cite{robertson2020excessive}, there is a growing concern about the potential misuse of such data, including unauthorized resale, identity theft, and other forms of exploitation. This necessitates the implementation of robust data governance frameworks that comply with privacy regulations such as the General Data Protection Regulation (GDPR)~\cite{EuropeanParliament2016a} and incorporate advanced cryptographic and anonymization techniques to safeguard consumer privacy. GAN models have been used to generate synthetic data, ensuring privacy-preserving downstream tasks~\cite{fonseca2023tabular}. The challenge lies in balancing the utility of the data for analytics purposes with the imperative to protect consumer privacy, a concern we aim to address in our research.

In light of these complexities and rapidly increasing data volumes, there is a pressing need for advanced research in deep learning for assortment management. While retail transactions provide valuable insights, they merely scratch the surface of the myriad of possible transaction combinations. Delving deeper into this expansive realm of potential transactions can reveal pivotal insight into aspects like product demand trends, shifting consumer inclinations, pricing dynamics, and variations tied to seasons in a previously undiscovered way. These data, encompassing transactional details, customer demographics, and product specifications, empower retailers to enhance customer service. Deep learning, with its ability to process vast amounts of data and discern patterns that might be imperceptible to traditional algorithms, offers a promising avenue for tackling the multifaceted challenges of assortment planning. The sheer volume and granularity of retail data and the dynamic nature of consumer preferences make it a prime candidate for applying deep learning techniques. By leveraging deep learning, retailers can potentially achieve more accurate demand forecasts~\cite{salinas2020deepar}, optimize inventory levels in real-time, and tailor assortments to individual consumer preferences, thereby driving increased sales and profitability~\cite{fisher2006retail}.

We believe our contributions -- delivering a framework of transactional data generation under stock constraints -- unlock additional opportunities in \textbf{assortment optimization} and \textbf{dynamic consumer behavior modeling}. 
    Existing methodologies~\cite{goyal2016near, sinha2017optimizing} underline the intricate nature of the assortment optimization problem. Its depth, focusing on the product mix within a category, encapsulates the challenge of balancing the introduction of new products against the potential cannibalization of sales from existing products~\cite{mccoll2020estimating}.
    Also, the sequential nature of purchases requires consideration of evolving behaviors and enduring patterns, adding layers of intricacy to modeling. Legacy approaches applied in traditional retail cannot ensure the agility to address the pace of changing preferences~\cite{hollebeek2019customer}.
Moreover, despite the extensive volume of transactional data, individual customer interactions with the vast array of products and broad assortments often result in sparse datasets~\cite{apeh2011customer}. It makes many related tasks computationally intractable, highlighting the need for more robust modeling methodologies.

\subsection{Objectives and Contributions} \label{objectives}
Our innovative methodology authentically simulates sequences of customer product baskets over time, leveraging a comprehensive dataset of customer transactions. Our approach can simulate plausible individual customer transactions, offering a holistic view of buyers' behaviors. This capability paves the way for novel applications, from predicting future purchases to facilitating research access to simulated datasets that mirror restricted original data due to privacy constraints. We hypothesize that this solution can be beneficial in estimating demand transference and its incrementality -- we discuss these notions in more detail in Section~\ref{conslusion}.

The primary objective of our research and the described study is to advance the field of customer modeling by introducing innovative techniques for generating realistic retail transactions under stock availability limitations. We aim to bridge the gaps between traditional methods, ongoing research, and the evolving needs of the retail sector, ensuring that simulated data is authentic and reflective of real-world consumer behaviors. Our key contributions are as follows:
\begin{outline}
    \1 \textbf{Enhanced Product and Consumer Representation:} We introduce an innovative approach of combining SKUs with transactional data, performing a comparative analysis of product embeddings using both the classic word2vec~\cite{mikolov2013efficient} method and the advanced hyper-graph model, Cleora~\cite{rychalska2021cleora}. Similarly, our research contrasts consumer embeddings generated through a recurrent neural network (RNN)~\cite{sherstinsky2020fundamentals} with those produced by Cleora, offering a comprehensive perspective on representation efficacy.
    \1 \textbf{Innovative Use of GAN with Stocks Data:} Combined with stock data availability, our adoption of the classic GAN architecture for generative modeling pushes the research in the field forward, ensuring that the generated data closely mirrors real-world transactional patterns. To the best of our knowledge, variations of stock data representation (weighted and unweighted embeddings) have been tested for the first time in our research to uncover the impact on generated transaction accuracy in the frame of the GAN architecture.
\end{outline}

Beyond theoretical contributions, our research has tangible, practical implications for the retail sector. By generating authentic transactional data, retailers can gain actionable insights, optimize inventory management, and enhance customer experiences.

\section{Related Work} \label{related}
While GANs were initially popularized for image generation tasks, their applications have expanded across various domains. For instance, they have been used for super-resolution tasks to generate photo-realistic images at large upscaling factors~\cite{ledig2017photo}. In feature learning, Context Encoders utilize GANs for inpainting, where the network learns to fill in missing parts of images conditioned on their surroundings~\cite{pathak2016context}. Moreover, GANs have been employed for domain adaptation tasks, effectively bridging the domain gap between different data distributions~\cite{tzeng2017adversarial}. GANs have also been successfully employed in various medical~\cite{zhou2023gan} and financial~\cite{assefa2020generating,takahashi2019modeling} applications -- extending beyond their traditional visual data domain. 

To our knowledge, there are two most relevant and practical applications of GANs in online~\cite{kumar2018ecommercegan} and offline~\cite{doan2018generating} retail. Both studies use product and customer embeddings to capture the semantic and sequential relationships between products and customers. Product embeddings were derived from textual descriptions of products using a pre-trained language model, while customer embeddings were learned from transactional data using an autoencoder. The authors then trained a conditional GAN to generate a basket of products for a given customer and week and an LSTM to update the customer's state after each basket. This way, they could sample over a distribution of a customer's future sequence of baskets, which could vary depending on the customer's preferences, needs, and behavior. The authors evaluated their method on a real-world retail dataset. They showed it could produce baskets similar to real baskets in product features, basket sizes, and sequential patterns. They also showed that the generated baskets were hard to distinguish from the real baskets by a classifier, indicating that their method could generate realistic and diverse sequences of transactions.

Our work extends the above approaches by incorporating stock data availability into the generation process. Stock data availability refers to the information about the quantity of products in the store or warehouse, which can affect the customer's purchase decisions and the retailer's operations planning. We acknowledge the scarcity, quality, and availability challenges of these data among retailers; however, ignoring stock data availability can lead to unrealistic or suboptimal sequences of transactions, as customers may not be able to buy the products they want or need, or retailers may not be able to replenish the products in time. Therefore, given that we possess a rare proprietary dataset with detailed stock information through a commercial partnership, we propose to modify the conditional GAN to consider the stock data availability when generating and updating the baskets. We hypothesize that this will result in more accurate and valuable sequences of transactions for retail analytics. Another more broadly related study addresses data generation aspects introducing CTAB-GAN~\cite{zhao2021ctab} for better privacy preservability. Our research echoes the considerations described by authors with a more specific narrowed retail industry application.

\section{Proposed Approach} \label{method}

 The methodology presented in this section aims to bridge the gap between theoretical advancements in machine learning and practical challenges in offline retail analytics. We employ Generative Adversarial Networks (GANs) as a foundational tool for simulating customer behavior, extending their capabilities to accommodate real-world constraints such as SKU stock limitations.

\subsection{Overview of Generative Adversarial Networks} \label{gans}

Generative Adversarial Networks (GANs)~\cite{goodfellow2014generative} are acknowledged for their ability to generate data closely resembling real-world distributions. The networks consist of two neural networks, the Generator and the Discriminator, trained in a min-max adversarial setting. The Generator aims to produce synthetic data, while the Discriminator's role is to distinguish between real and generated data. The training process involves a game-theoretic approach where both networks are optimized until the Generator produces data that the Discriminator can hardly distinguish from real data.

Further developed, Wasserstein GAN (WGAN) and its modification with gradient penalty (WGAN-GP) address the issue of training instability commonly observed in standard GANs by adopting the Wasserstein distance as the loss function. This modification leads to more stable training and higher-quality generated samples~\cite{gulrajani2017improved}.

\begin{equation}
\min_{G} \max_{D} \mathbb{E}_{x \sim p_{\text{data}}}[\log D(x)] + \mathbb{E}_{z \sim p_{z}}[\log(1 - D(G(z)))]
\end{equation}
where we adopt the following notation (also in all our equations):
    G -- the Generator network, responsible for generating synthetic data that aims to mimic the real data distribution;
    D -- the Discriminator network, responsible for distinguishing between real and synthetic (generated) data;
    x -- a sample from the real data distribution; 
    z -- a sample from the latent space or noise distribution.

\textbf{Wasserstein GAN (WGAN):} A significant advancement in the GAN landscape is the introduction of Wasserstein GAN (WGAN). WGANs address the issue of training instability commonly observed in standard GANs by adopting the Wasserstein distance as the loss function. This modification leads to more stable training and higher-quality generated samples~\cite{gulrajani2017improved}.
\begin{equation}
\min_{G} \max_{D} \mathbb{E}_{x \sim p_{\text{data}}} [D(x)] - \mathbb{E}_{z \sim p_{z}} [D(G(z))]
\end{equation}
In WGAN, the Discriminator (often referred to as the "Critic" in this context) is not restricted to output probabilities, allowing for a broader range of output values.

\textbf{WGAN with Gradient Penalty (WGAN-GP):} Building upon WGAN, the WGAN-GP introduces a gradient penalty term to the loss function. This addition further stabilizes the training process and enables the use of more complex architectures without the need for extensive hyperparameter tuning~\cite{gulrajani2017improved}.

\begin{equation}
\min_{G} \max_{D} \mathbb{E}_{x \sim p_{\text{data}}} [D(x)] - \mathbb{E}_{z \sim p_{z}} [D(G(z))] + \lambda \mathbb{E}_{\hat{x} \sim p_{\hat{x}}} [(||\nabla_{\hat{x}} D(\hat{x})||_2 - 1)^2]
\end{equation}
where $\lambda$ is the penalty coefficient and $\hat{x}$ is sampled uniformly along a straight line between a pair of real and generated data points.

Unlike the above classical architectures, we leverage the conditional and controllable GAN approach demonstrated in~Section~\ref{design}, integrating a stock representations. Our target objective function used to establish  generative adversarial network looks as follows:
\begin{equation}
\min_{G} \max_{D} \left( \mathbb{E}_{x \sim p_{\text{data}}} [D(x)] - \mathbb{E}_{p, s} [D(G(p, s))] + \lambda \mathbb{E}_{\hat{x} \sim p_{\hat{x}}} [(||\nabla_{\hat{x}} D(\hat{x})||_2 - 1)^2] \right)
\end{equation}

\noindent where we adopt the following notation:
\begin{itemize}
    \item 
    $G$ -- the Generator network, responsible for generating synthetic data that aims to mimic the real data distribution;
    \item 
    $D$ -- the Discriminator network, responsible for distinguishing between real and synthetic (generated) data;
    \item
    $x$ -- a sample from the real data distribution; 
    $p$ -- product embedding, $s$ -- stock embedding,  $\lambda$ -- the penalty coefficient, and $\hat{x}$ -- sampled uniformly along a straight line between a pair of real and generated data points.
    \end{itemize}
In this way, we secure the possibility of simulation of various stock scenarios that are crucial for retailers' category planning and assortment management decisions. For example, when a retailer removes a product from the assortment, it will result in a different stock embedding -- a different set of transactions will be generated by the network under new stock conditions.

\subsection{Features and Representations} \label{features}

The proposed GAN model leverages multiple types of embeddings and features, including stock, customer, and product embeddings. We aim to compare different ways of representing products, customers, and stocks to observe changes in model performance and uncover the most efficient combination. Therefore, we engineer the following features and representations:
\begin{itemize}
    \item \textbf{Product:} Similarly to the related research we refer to in Section~\ref{related}, we generate a simple representation using word2vec~\cite{mikolov2013efficient}, but also leverage Cleora~\cite{rychalska2021cleora} as a potential more advanced alternative that has proven to be effective in one of our previous study [removed for blind review]
    \, in a comparable setting. Unlike word2vec, we use Cleora to generate embeddings not only using product names but also including information about products' interactions within purchase baskets.
    \item \textbf{Customer:} We use RNN architecture that has been used by other researchers for similar applications~\cite{salampasis2021comparison} to extract consumer representation from the last layer of the network's hidden layer. As a comparison, we use the above-mentioned Cleora algorithm with relevant input parameters to enable both product and consumer embedding outputs.
    \item \textbf{Dates:} Retail data often exhibit strong seasonal patterns. By using cyclic features for dates, your model can capture these patterns effectively, allowing for more accurate and realistic transaction generation. Hence, we extract cyclic features from the transaction dates, such as the day of the week, the day of the month, and the month itself.
    \item  \textbf{Price:} We transform unit price using a natural logarithm to reduce the dynamic range of a variable since some values are significantly larger than others. Variability of prices is expected when handling a wide range of assortment in large-scale retail.
    \item \textbf{Stocks:} The generation of stock embeddings is a two-step process involving unweighted and weighted aggregation of product embeddings (we use Cleora-generated embeddings that hold more information than the ones generated with word2vec). Firstly, for each unique combination of site and date, a subset of product indices with corresponding product embeddings and quantities is extracted. For unweighted representation, the mean of the product embeddings is computed, whereas for weighted - a weighted average of the product embeddings is computed using the quantities as weights.
\end{itemize}

Lengths of product, customer, dates, and price are 1024, 256, 6, and 1, respectively; this results in a 1287-length transnational item representation. Since we use products' embeddings to compute stock representation, it has the same length - 1024. All features and representations are scaled to the range $[-1, 1]$.
\begin{figure}[h]
\centering
\includegraphics[width=0.8\textwidth]{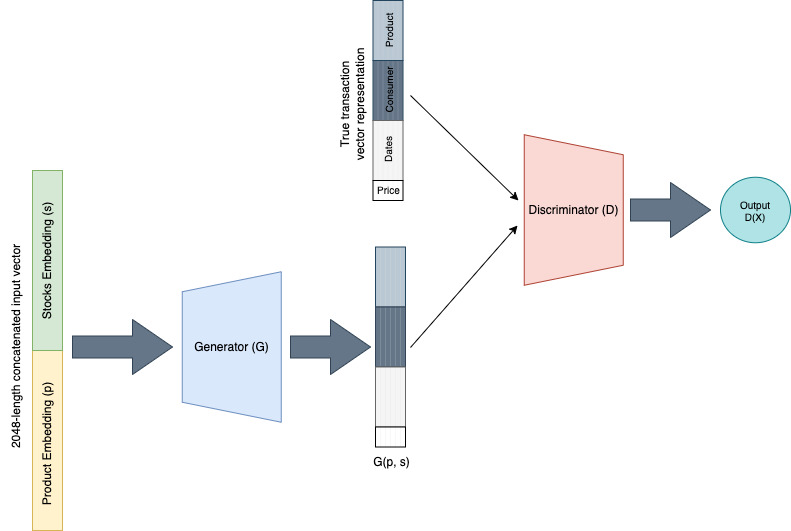}
\caption{Proposed augmentation of conditional GAN.}
\label{fig:architecture}
\end{figure}

\section{Experiments and Results} \label{results}

The objective of this section is to rigorously evaluate the efficacy of the proposed GAN model in generating realistic transaction data that adheres to real-world constraints. We employ a comprehensive experimental setup, leveraging proprietary data and employing multiple performance metrics to assess the quality of generated transactions.

\subsection{Dataset} \label{dataset}
The dataset used in this research is provided by one of the largest retailers in Europe, holding a significant share of the domestic market. This unique dataset offers a rare opportunity to study consumer behavior and retail operations at scale. The data encompasses various aspects of retail transactions, including but not limited to Customer, Product (Product ID, Product Name, Size, and other metadata), Transactional details (e.g., ID, price, quantity, date, etc.), Store (Store ID, City, and other metadata), Stocks (availability and quantity of a product per day per store).

The dataset is particularly valuable for several reasons. Its scale allows for robust statistical analysis and the training of complex machine learning models like GANs. The diversity means it is likely to be representative of broader shopping behaviors in Europe, thereby increasing the generalizability of the research findings. The dataset includes fine-grained details such as product embeddings, transaction dates, and unit prices, which are crucial for the nuanced understanding and modeling of consumer behavior. Sourced from an industry leader, the dataset reflects real-world retail operations, making the research findings directly applicable to practical challenges in retail management.

The data offers a broad temporal and spatial scope as it was collected from 986 distinct retail sites over 58 days or approximately eight weeks. It encompasses an impressive 2,061,078 customers. The dataset is particularly rich in transactional data, containing 31,183,932 transactions. It records an average of approximately 3,464,881 transactions per week. This high volume of transactional data is crucial for training sophisticated machine learning models, such as the Generative Adversarial Network (GAN) model proposed in this study. Each transaction involves an average of nearly 14 products, of which around 11 are unique. This diversity is further emphasized by the fact that each customer purchases an average of approximately 106 unique products, making the dataset highly suitable for studying assortment planning. About half (1,049,775) of the repeated customers in our training dataset have engaged in transactions at multiple sites. This high customer loyalty indicates recurring purchase patterns essential for accurate demand forecasting. Moreover, the dataset includes 29,393,436 transactions involving more than two products and 2,055,421 transactions involving at least five different products in a single transaction, while the overall assortment of the retailer contains more than 55,000 unique SKUs. These statistics underscore the complexity of consumer buying behavior, which is a central focus of this research.

In summary, the dataset's extensive scale, diversity, and granularity make it an exceptional resource for investigating the challenges and opportunities in retail operations, particularly in assortment planning and demand forecasting. Given its real-world relevance and the scale at which the data has been collected, the research findings are expected to have direct and significant implications for the retail industry, especially for large-scale retailers.

\subsection{Model Design and Training}\label{design}

We experimented with various architectures and configurations of our GAN model. There are two worth noting observations based solely on the architectural experimentation: WGAN-GP 
significantly increases the stability of training even in the first epochs.
At the same time, slightly different architectures for Generator and Discriminator help to avoid mode collapse at the initial training stages (even when a gradient penalty is not applied). The final architecture (Fig.~\ref{fig:architecture}) we have used in the frame of this study is described below.
\begin{figure}[h]
\includegraphics[width=\textwidth]{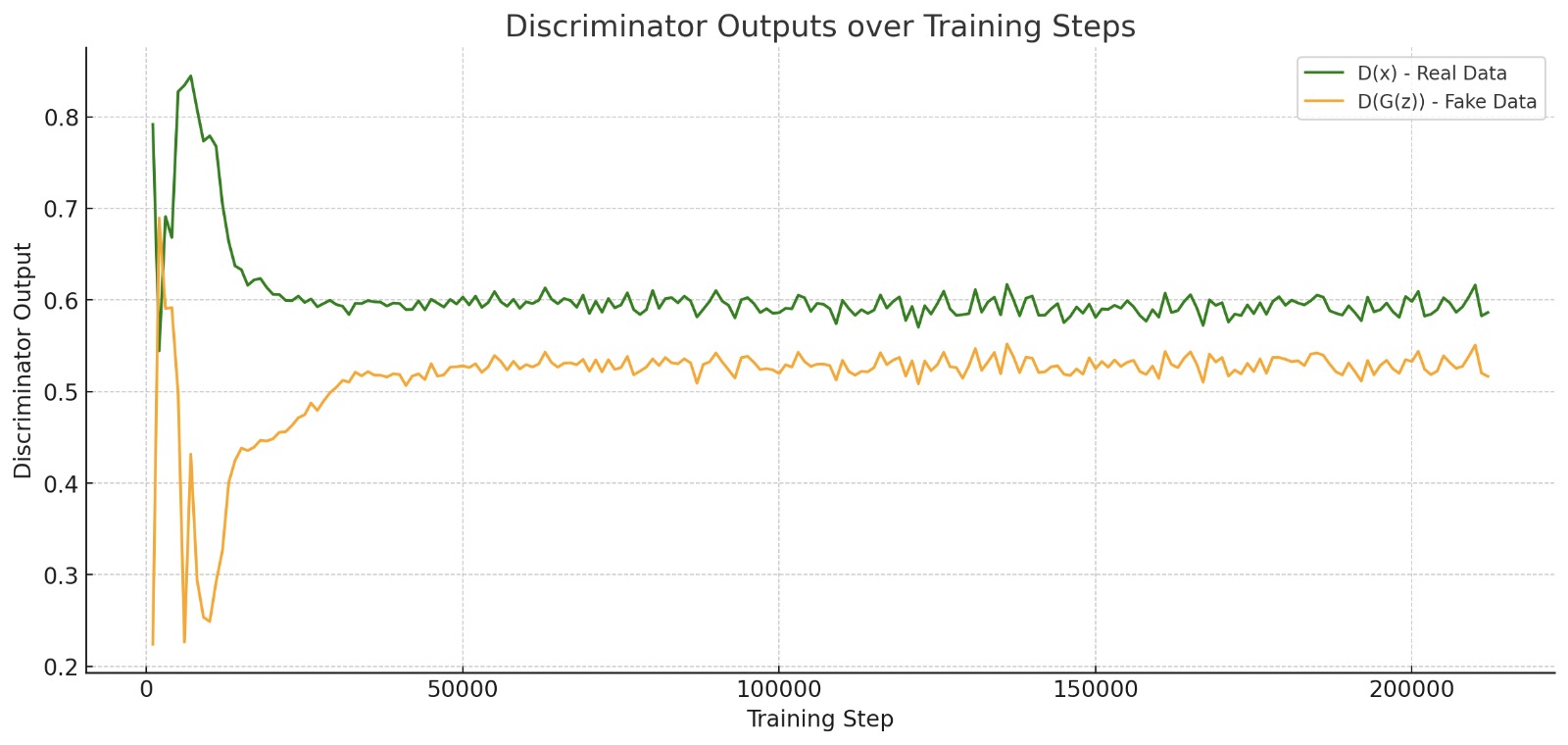}
\caption{Stabilization of discriminator training process (illustrative example).}
\centering
\label{fig:training}
\end{figure}

The Generator is a neural network comprising four fully connected layers with the following dimensions:
\begin{outline}
    \1  \textbf{Layer I}: 2048 input units (product and stock embeddings) to 1024 output units, followed by a LeakyReLU\footnote{https://pytorch.org/docs/stable/generated/torch.nn.LeakyReLU.html} activation.
    \1 \textbf{Layer II}: 1024 to 512 units, followed by a LeakyReLU activation.
    \1 \textbf{Layer III}: 512 to 256 units, followed by a LeakyReLU activation.
    \1 \textbf{Output Layer}: 256 to 1287 units, followed by a Tanh activation.
\end{outline}

The Discriminator also consists of a neural network with three fully connected layers:
\begin{outline}
    \1 \textbf{Layer I}: 1287 input units to 512 output units, followed by a LeakyReLU activation and a dropout layer.
    \1 \textbf{Layer II}: 512 to 128 units, followed by a LeakyReLU activation and another dropout layer.
    \1 \textbf{Output Layer}: 128 to 1 unit, followed by a sigmoid activation.
\end{outline}

The generator's loss function is a weighted combination of the GAN loss and a reconstruction loss (Equation~\ref{recloss}). The GAN loss aims to deceive the discriminator, while the reconstruction loss ensures that the generated embeddings are close to the real product embeddings.
\begin{equation} \label{recloss}
\mathcal{L}_{\text{R}} = \frac{1}{N} \sum_{i=1}^{N} \left\| \mathbf{e}_{\text{real}, i} - \mathbf{e}_{\text{gen}, i} \right\|^2_2
\end{equation}
where
    \( N \) -- number of samples, \( \mathbf{e}_{\text{real}, i} \) -- real product embedding for the \( i^{th} \) sample, \( \mathbf{e}_{\text{gen}, i} \) -- generated product embedding for the \( i^{th} \) sample.

Two different optimization algorithms are employed for training the Generator and Discriminator:
    the Generator uses the Adam optimizer~\cite{kingma2014adam} with a learning rate of 0.0002 and \( \beta \) values of (0.5, 0.999), and the
    Discriminator uses the RMSprop optimizer~\cite{tieleman2012lecture} with a learning rate 0.0002.
The Discriminator is trained five times for each Generator update. The Discriminator loss is computed using real and fake orders and a gradient penalty term to enforce the Lipschitz constraint~\cite{liu2020lipschitz}.

The training process of the proposed Generative Adversarial Network (GAN) model is formalized as Algorithm~\ref{alg:gantraining} to provide a structured and easily interpretable overview of the vital computational steps. Although we cannot publish a full PyTorch implementation because of the proprietary nature of collaboration and data sourcing, by delineating the process in this manner, we aim to offer a transparent and replicable framework that can be readily understood and implemented by researchers and practitioners alike. Training, evaluation, and inference processes were performed using Nvidia DGX A100 40GB.

\begin{algorithm}
\caption{Training Process of the Proposed GAN Model} \label{alg:gantraining}
\begin{algorithmic}[1]
\State \textbf{Input:} Training dataset, batch size, number of epochs
\State \textbf{Initialize:} Generator $G$, Discriminator $D$
\State \textbf{Initialize:} Optimizers $optimizer\_G$, $optimizer\_D$

\For{epoch in 1, 2, ..., epochs}
    \For{mini-batch in DataLoader}
        \State Extract real orders and stock embeddings
        \State Extract product embeddings from real orders
        \State Move all data to computation device (GPU)
        
        \For{iteration in 1, 2, ..., $n_{\text{critic}}$}
            \State Generate fake orders using $G$
            \State Compute Discriminator loss $\mathcal{L}_{\text{D}}$
            \State Update $D$ using $optimizer\_D$
        \EndFor
        
        \State Generate fake orders using $G$
        \State Compute Generator loss $\mathcal{L}_{\text{G}}$
        \State Update $G$ using $optimizer\_G$
    \EndFor
\EndFor
\end{algorithmic}
\end{algorithm}

\subsection{Performance Metrics}
Evaluating the performance of Generative Adversarial Networks (GANs) is challenging, especially in non-visual domains. The limitations of GANs in discriminative tasks and their sensitivity to domain shifts make the evaluation process intricate~\cite{saxena2021generative}. To address these challenges, we employ a multi-faceted approach to assess the quality of the generated transactions using the following metrics.

\noindent \textbf{Jensen-Shannon Divergence (JSD):} Symmetric measure of the similarity between two probability distributions. It is particularly useful for comparing the distributions of real and generated transactions in our context. The JSD is the average of two Kullback-Leibler divergences, one for \textit{P} and one for \textit{Q}, the probability distributions of real and generated transactions, respectively.
    \begin{equation}
    \text{JSD}(P, Q) = \frac{1}{2} D_{KL}(P \parallel M) + \frac{1}{2} D_{KL}(Q \parallel M)
    \end{equation}
    where
    \begin{equation}
    M = \frac{1}{2}(P + Q),
    D_{KL}(P \parallel Q) = \sum_{x} P(x) \log \frac{P(x)}{Q(x)}
    \end{equation}
    In retail transactions, the tail behavior of the distribution can be crucial. JSD is sensitive to differences in the tails of the distributions, making it a suitable metric for our application. A lower JSD value means that the two distributions are more similar in an information-theoretic sense.
    
    \noindent \textbf{Earth Mover's Distance (EMD):} Measures the distance between the probability distributions of real and generated transactions. Similarly to the above, lower EMD value indicates that less "work" is required to transform one distribution into another, suggesting that the two distributions are more similar.
    \begin{equation}
    \text{EMD}(P, Q) = \inf_{\gamma \in \Pi(P, Q)} \int_{X \times Y} d(x, y) d\gamma(x, y)
    \end{equation}
    
    \noindent \textbf{Classification Accuracy (Acc):} Utilizes a simple binary classifier to distinguish between real and fake samples in the downstream task. We follow the same approach used previously in~\cite{doan2018generating} to create a classifier different from the one used to train GAN. Accuracy around 0.5 would mean the classifier cannot distinguish real transaction representation from a generated one.

By employing these metrics, we aim to provide a comprehensive and robust evaluation of the quality of transactions generated by our GAN model. Each metric offers a unique perspective: EMD focuses on the overall distribution, Accuracy 
provides an operational view, and JSD gives a balanced and bounded measure. It's worth highlighting that both EMD and JSD are used to measure the similarity between two probability distributions. Still, they have different properties and sensitivities that make them suitable for different types of analyses. EMD is useful for understanding the overall shape and spread of the distribution, including the impact of outliers, which could represent high-value transactions, while JSD compares the core behaviors of customer transactions, especially when outliers are less of a concern. Together, they allow for a nuanced understanding of the model's performance.

\subsection{Results and Discussion}
In this section, we present the results of our experiments designed to evaluate the efficacy of various GAN models in generating plausible retail transactions. The models were trained using different combinations of consumer and product embeddings, with and without including stock information, including its weighting variation.

\begin{table}[h]
\centering
\caption{Results for Models Trained on Various Embeddings}
\label{table:results}
\begin{tabular}{ccccccccccc}
\hline
\multirow{2}{*}{Consumer} & \multirow{2}{*}{Product} & \multicolumn{3}{c|}{Stocks (Unweighted)} & \multicolumn{3}{c|}{Stocks (Weighted)} & \multicolumn{3}{c}{No Stocks} \\
\cline{3-11}
& & EMD & JSD & Acc & EMD & JSD & Acc & EMD & JSD & Acc \\
\hline
Cleora & Cleora & 0.23 & 0.12 & 0.60 & \textbf{0.18} & \textbf{0.09} & \textbf{0.58} & 0.35 & 0.20 & 0.68 \\
Cleora & w2v & 0.28 & 0.16 & 0.63 & 0.24 & 0.12 & 0.61 & 0.40 & 0.24 & 0.71 \\
RNN & Cleora & 0.21 & 0.10 & 0.58 & \textbf{0.16} & \textbf{0.07} & \textbf{0.54} & 0.33 & 0.18 & 0.66 \\
RNN & w2v & 0.26 & 0.14 & 0.61 & 0.22 & 0.11 & 0.59 & 0.38 & 0.22 & 0.69 \\
\hline
\end{tabular}
\end{table}

In evaluating our GAN models, summarized in Tab.~\ref{table:results}, we observed significant performance variations based on the type of embeddings and the inclusion of stock information. The EMD and JSD metrics were lower for models trained with weighted stocks, indicating a closer match to the real data distribution. Particularly, models utilizing RNN for consumer embeddings and Cleora for product embeddings achieved the best results with EMD and JSD values of 0.16 and 0.07, respectively. These models also achieved a classification accuracy closest to 0.5, which is considered ideal in the context of GANs as it signifies that the generated transactions are nearly indistinguishable from the real ones.

When comparing these results to models trained without stocks, the importance of incorporating stock information becomes evident. Models without stock information had higher EMD and JSD values, indicating a greater divergence from the real data distribution. Moreover, their classification accuracy was far from the ideal 0.5 mark, making them more easily distinguishable from real transactions. These findings strongly support our hypothesis that weighted stocks provide more informative embeddings, thereby enhancing the performance of GANs in generating realistic retail transactions.

\section{Conclusion and Future Work} \label{conslusion}
In this study, we have presented a novel approach to generating realistic retail transactions using GANs. Our work is distinguished by incorporating stock information into the model, a feature often overlooked in previous research. Through the evaluation involving multiple metrics (i.e. EMD, JSD, Acc), we demonstrated that including weighted stock information significantly enhances the quality of the generated transactions. Our research contributes to the growing body of work on applying deep learning techniques in the retail industry, offering a new avenue for enhancing the realism and utility of generated transaction data.

The current research serves as a foundational step for employing deep learning in assortment optimization within large-scale retail. Our findings pave the way for several future research directions. One immediate extension is the development of a transferable demand model that can adapt to new data quickly, offering a scalable solution for large retailers' category management efforts~\cite{karampatsa2017retail}. While the notion of transferable demand was studied earlier with a statistical approach~\cite{mahalanobish2017capturing}, our goal is to create a more robust deep learning model that not only adapts to changing customer preferences but also can be easily transferred to different retail settings or scaled up to accommodate larger datasets. This would involve incorporating more advanced machine learning techniques and perhaps integrating other forms of data to create a more holistic model. The potential of the GAN model in this study also suggests that other neural network architectures could be explored for similar applications. Finally, the ultimate validation of these models would come from their deployment in real-world retail settings. By focusing on these areas, future work can offer more robust and effective solutions for assortment optimization, contributing to the broader application of deep learning and computer science in complex industrial problems.

\section*{Acknowledgements}
This work was supported by computational grant received from the Warsaw University of Technology allowing to perform modeling using HPC cluster.

\FloatBarrier

\bibliographystyle{unsrtnat}
\bibliography{bibliography}  

\end{document}